\documentclass{article}

\usepackage[preprint,nonatbib]{neurips_2021}

\usepackage[colorlinks]{hyperref}
\hypersetup{
    colorlinks=true,
    linkcolor=blue,
    citecolor=blue,
    filecolor=blue,
    urlcolor=blue
}
\usepackage{url}            
\usepackage{booktabs}       
\usepackage{amsfonts}       
\usepackage{nicefrac}       
\usepackage{microtype}      
\usepackage{cite}
\usepackage{wrapfig}
\usepackage{enumitem}
\usepackage{xcolor}         

\usepackage{arydshln}
\usepackage{epsfig}
\usepackage{graphicx}
\usepackage{amsmath,amssymb,bm}
\usepackage{multirow}
\usepackage{multicol}
\usepackage[ruled,vlined]{algorithm2e}
\SetKwInput{KwInput}{Input}
\SetKwInput{KwReturn}{Return}

\newcommand{\eg}{\emph{e.g.}}
\newcommand{\ie}{\emph{i.e.}}

\newcommand{\ee}[2]{${#1}_{{#2}}$}
\newcommand{\eeb}[2]{$\mathbf{{#1}_{{#2}}}$}
\newcommand{\et}[2]{${#1}_{\pm{#2}}$}
\newcommand{\etb}[2]{$\mathbf{{#1}_{\pm{#2}}}$}

\def\vo{{\bm{o}}}

\def\vz{{\bm{z}}}

\title{Divide and Rule: Recurrent Partitioned Network for Dynamic Processes}

\author{
  Qianyu Feng \\
  ReLER Lab\\
  University of Technology Sydney\\
   \And
   Bang Zhang \\
   DAMO Academy\\
   Alibaba Group\\
   \AND
   Yi Yang \\
   Zhejiang University \\
}

\begin{document}

\maketitle

\begin{abstract}
In general, many dynamic processes are involved with interacting variables, from physical systems to sociological analysis. The interplay of components in the system can give rise to confounding dynamic behavior. Many approaches model temporal sequences holistically ignoring the internal interaction which are impotent in capturing the protogenic actuation. Differently, our goal is to represent a system with a part-whole hierarchy and discover the implied dependencies among intra-system variables: inferring the interactions that possess causal effects on the sub-system behavior with \textsc{REcurrent partItioned Network} (REIN). The proposed architecture consists of (i) a perceptive module that extracts a hierarchical and temporally consistent representation of the observation at multiple levels, (ii) a deductive module for determining the relational connection between neurons at each level, and (iii) a statistical module that can predict the future by conditioning on the temporal distributional estimation. Our model is demonstrated to be effective in identifying the componential interactions with limited observation and stable in long-term future predictions experimented with diverse physical systems. 
\end{abstract}

\section{Introduction}

Dynamics simulation involves multiple bodies are widely adopted in engineering fields of research, especially in physics, robotics~\cite{khatib2009robotics,vzlajpah2008simulation,young1988variable} and biotechnology~\cite{ellner2011dynamic,villaverde2016structural}.
In many practical applications, the complexity for modeling the dynamics emerges within the combination and interaction of sub-modules in physical systems. If we want to enable the agent to simulate the real process, it is a prerequisite to figuring out the underlying functions for controlling the system dynamically. Some systems may be adequately described by deterministic equations. However, it will be intractable if the system possesses observational noise, unobserved factors, or intrinsic randomness. 

Under the dynamic settings, we consider a physical system as the composition of constituents varying over time. As an important property, multi-body system formalisms usually offer an algorithmic, computer-aided way to model, analyze, simulate and optimize the arbitrary motion of possibly large amount of inter-connected bodies. 
So, \textit{how do models succeed in capturing the dynamic patterns?} 
In a system with general purpose, it is required to perform specific functionalities or behavior. 
It is straightforward that the system could be structured with a graph network with correspondence to its physical connections. Each part of the system can be divided into several sub-parts accordingly. This facilitates the components to be controlled in a hierarchical way. In addition, it is important to keep the connections dynamic which requires the network to infer the relationship between variables on the fly.
Rather than allocating neurons to represent the system parts with a parse tree and assigning pointers to its ancestor and descendants, we consider using an appropriate vector to exchange information with the relatives. Formed in a hierarchical structure, we propose \textsc{REcurrent partItived Network} (REIN) consisting of stacks of auto-encoders learning hierarchical representations. Take the human skeleton for example, each joint represents the lowest level of the system. At a higher level, the limbs and trunk are composed of different groups of joints. 

In this paper, we are primarily concerned with modeling dynamic processes with interacting variables based on the observation. The system can be compartmented into different levels of sub-systems. 
\textit{What are the underlying formational mechanism of variables in the interacting system?}
There are mainly two types of causality considered in the context of dynamic sequences. The most common is Granger causality~\cite{granger1980testing} which has been widely applied in the industry. However, it often fails in the presence of contemporaneous effects, which has motivated the development of dynamic-causality~\cite{ye2015distinguishing,Sugihara496}.
To recover a dynamic causal model from observation, we need to first learn a compact state representation of sub-systems and then infer a causal graph among the variables to identify hidden confounders. In order to learn the dynamic mechanism in an end-to-end style, we target at simulating data samples with different causality but with a shared causal inference module.

To sum up, our main contribution exists in proposing a hierarchical recurrent architecture, \textsc{REcurrent partItived Network} (REIN), for modeling the interplay of variables in dynamic processes. The proposed architecture consists of (i) a perceptive module that extracts a hierarchical and temporally consistent representation of the observation at multiple levels, (ii) a deductive module for determining the relational connection between neurons at each level, and (iii) a statistical module that can predict the future by conditioning on the temporal distributional estimation. 
Different scenarios and settings are considered in the evaluation, which provides various combinations of variables, \ie, data from unknown interventions on the underlying system.
Our model is demonstrated to be effective in identifying the interactions from limited observation and carry out long-term future predictions in diverse physical systems. 
In a range of experiments on physical simulations, we show that REIN possesses a favorable inductive bias that allows it to discover ground-truth physical interactions with high accuracy in a completely unsupervised way. We further show on real motion capture data that our model can accurately predict the dynamics many time steps into the future.

\section{Background and Related Work}

\textbf{Hierarchical representing the system with part-whole structure}
It is at almost no cost for people to parse the scene and object at different levels of integrity. We need to build a model that can understand the mechanism and manipulate data so that the model is able to reproduce the dynamic behavior from the observation. 
\begin{wrapfigure}{r}{0.6\textwidth}
    \centering
    \includegraphics[width=0.6\textwidth]{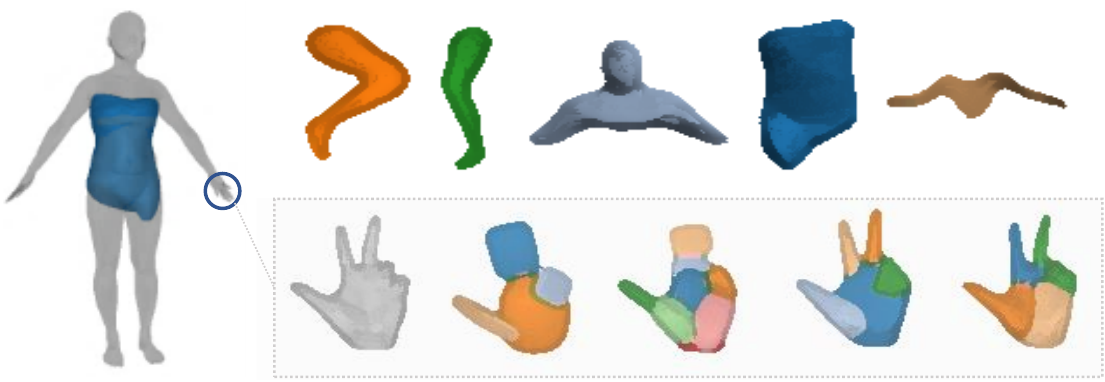}
    \caption{Example of the hierarchical part-whole structure in human skeleton.}
    \vspace{-2mm}
    \label{fig:teaser}
\end{wrapfigure}
In a computer that has general purpose and dynamic memory, the obvious way to represent the part-whole hierarchy~\cite{girju2003learning,winston1987taxonomy,bateman1997does} for a specific structure is to create a graph structure~\cite{wu2019graph,luo2020dynamic,hajiramezanali2019variational,zhang2017random,song2019session} by 
dynamically allocating pieces of the memory to the nodes in the graph and giving each node pointers to the nodes it is connected to. Combining this type of dynamically constructed graphs~\cite{goyal2020dyngraph2vec,seo2018structured} with neural network learning techniques has recently shown great promise, but if the whole computer is a neural network, it is far less obvious how to represent hierarchies that are different for every sample if we want the structure of the neural net to be identical for all samples.


\textbf{Causal effects in dynamic interacting processes}
There exist several typical approaches to causal structural discovery: score-based, constraint-based, methods leveraging the structural asymmetries and those exploiting various intervention mechanisms. From a perspective, methods can be categorized as local, whereby edges are tested one at a time, or global, whereby an entire graph candidate is tested. 
It is only applicable if separability holds. Separability refers to the independence of the variables in the absence of causal interactions. Unfortunately, this is rarely the case in dynamic systems, where the current state of a variable may be heavily determined by the past of another. 
The authors of CausalVAE~\cite{yang2020causalvae} state that whilst many disentangled representation learning methods assume independence between latent factors~\cite{higgins2016beta,kyono2020castle,locatello2019challenging}, most latent factors behind real-world phenomena exhibit causal dependencies. They propose the use of a Variational Auto-Encoder~\cite{rezende2014stochastic,kumar2017variational}. The latent space of a VAE is usually parameterized by a set of exogenous factors. CausalVAE integrates a Causal Layer which transforms these exogenous latent factors into endogenous factors which reflect the causal semantics of the data.

\section{REIN: Recurrent Partitioned Network for Dynamic Process}

In this paper, we propose \textsc{REcurrent partItived Network} (REIN), for modeling the interacting variables of dynamic processes in a hierarchical way. In REIN, each neuron represents the states of a sub-system. The system can be first divided into multiple levels and constituents at each level can be learned with like-minded embeddings. In this way, the process can be modeled dynamically to capture the dynamics in the observation.

\subsection{Hierarchical Part-whole Partitioning}

\begin{figure}[t]
    \centering
    \includegraphics[width=\hsize]{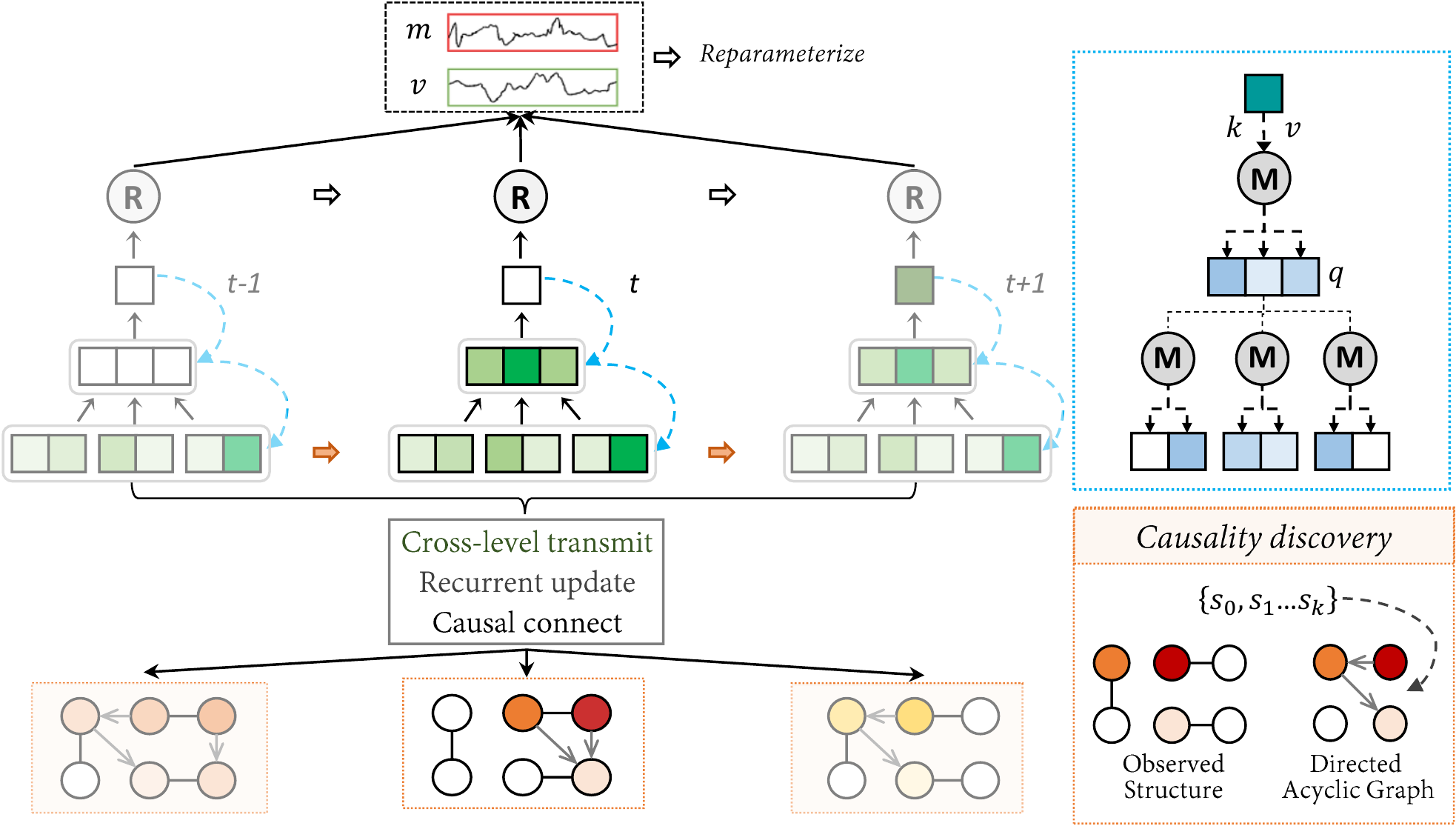}
    \vspace{3mm}
    \caption{Architecture of Recurrent Partitioned Network mainly consists of embeddings from : (i) \textit{upward direction}: states aggregated from the embeddings of the level below; 
    (ii) \textit{downward direction}: messages propagated from the embeddings of the level above; 
    (iii) \textit{sibling communication}: remnant states from the previous time steps at the same level. $\mathbf{R}$ represents for the recurrent cell and $\mathbf{M}$ denotes multi-headed attention layers. For reconstruction tasks, at each time step, the states of each neuron are further fed into a decoder to predict future steps.}
    \label{fig:architecture}
\end{figure}

Considering observation sampled from the measurable variables generated by a latent implicit function, we apply the partitioned neurons to recurrently model the dynamic states of each part. In a dynamic process, each data point is generated with unseen interventions on the underlying structure and parameters.
In this way, dynamics can be controlled part-to-part instead of generating as a whole. In general, the highest level represents the system as a whole while the lowest level represents each variable that can be identified as the smallest unit. 
The input control signal is fed into the global representation to control the system level-by-level and intervene in the dynamics of the lowest level. The generation of dynamics can be formed by passing the controlling messages in the top-down direction. Each neuron should be activated and updated when the input message is relevant to its embedding. The attention layers are used to activate the parts to be controlled. 

Figure~\ref{fig:architecture} depicts the architecture of our framework, which is composed of recurrent partitioned neurons for learning the representation hierarchically with information aggregated from \textit{top, bottom and peer} communication. 
The input vector contains a sub-vector as the current states and also the control signal. In addition, we also have decoders for reconstructing the dynamic process. 

The neurons in REIN can universally represent whatever takes place at the relevant location.
The bottom-up aggregation can be implemented with simple full-connection layers or GNNs. In the top-down direction, the attention mechanism is adopted to allocate representational and directional messages. Each lower level acts as a query to retrieve the relevant information from the level above. It is also explainable that each neuron represents a set of entities or elements. The embedding of each component is associated with a key and value from the upper level. In this way, each parent node can select the activated neurons to function on the fly and the best suited ones according to the attentive scores. So, at each time step, the embedding of each \textit{neuron} at a level can be updated with information from:
\begin{itemize}
    \item \textit{Upward direction}: local states that to be aggregated from the embeddings at the level below. The bottom-up modules extract valuable features from the lower-level neurons that get combined at the upper level;
    \item \textit{Downward direction}: messages delivered from the embeddings at the level above. The top-down modules generate different outputs for lower level neurons that might perform different dynamics; 
    \item \textit{Sibling communication}: remnant states propagated from the previous time steps at the same level, thus making reasonable and stable predictions over time.
\end{itemize}

\subsection{Peer propagation with causal effects}

Apart from relying to the physical structure, we aim to discover the invisible relationship among the neurons at the same level. For this purpose, we learn a Directional Acyclic Graph (DAG), $\mathbb{G}=( \mathcal{V}^{1:T}, \mathcal{E}^{1:T})$ for connecting the neurons, where $\mathcal{V}^{1:T}$ denotes the sub-components at different time steps. $\mathcal{E}^{1:T}$ is also learnable over time, denoting the varying relationships between the constituting components. Specifically, for each directed edge $(v_{m, i}, v_{m, j}) \in \mathcal{E}^{1:T}$, there are hidden confounders denoting the parameters of the relationship that determines the dynamical relationship and affects the behavior of the system. In a dynamic process, the edges among nodes represent the relationship among the changing variables which could crash and recover during the procedure. At level $m$, the bottom-up information can be aggregated with multiple implementations, \eg, MLPs, CNNs and GNNs, referred to as $\vz_u$. In the other direction, we utilize an attention mechanism that selects and then activates only a subset of the cells for each time step, denoted as $\vz_d$. 

Our goal is to perform the recovery of structural causal graph in a short sequence of data samples and simultaneously learn a shared dynamics model which makes counterfactual predictions into the future. 
${\mathcal{V}}^t = \{ \vo^t_{m, i} \}_{i=1}^{N} \in \mathbb{R}^2 $ represents the constituting components of level $m$ in the system at time step $t$.
Then, we use an inference module parameterized by $\phi$, that takes the sequence of components at level $m$ as input and predicts the edge set, 
    $\tilde{\mathcal{E}} = f_\phi(\tilde{\mathcal{V}}^{1:T}),$
$\tilde{\mathcal{V}}^{1:T}$ and $\tilde{\mathcal{E}}$ together constitute the discovered {\it causal effects}, conditioned on which, a regressive module, $f_\psi$, parameterized by $\psi$, aims to predict the state of the keypoints at time $t+1$,
    $\vz_c = f_\psi(\tilde{\mathcal{V}}^{1:T}, \tilde{\mathcal{E}}).$
We formalize our model as a Variational Auto-Encoder (VAE) to estimate the distribution.
In training phase, the latent embeddings at time step $t$ is calculated as:
\begin{equation}
    \begin{aligned}
    \mathbf{h}_t &= [\vz_u, \vz_d, \vz_c]|_t, \\
    \mu_\phi(t), \sigma_\phi(t) &= \mathrm{GRU}_\phi(\mathbf{h}_t),\\
    \vz_r &\sim \mathcal{N}(\mu_\phi(t), \sigma_\phi(t)),\\
    \end{aligned}
\end{equation}
$\mathbf{z}_r$ is a learnable variable re-parameterized from the learned Gaussian distribution $\mathcal{N}(\mu_\phi(t), \sigma_\phi(t))$. Then, together with the input control signal $\mathbf{a}$, \eg, the action label to be generated,
\begin{equation}
    \begin{aligned}
    \mathbf{g}_t &= \mathrm{E_n}(\vz_r, \mathbf{a}, \mathbf{h}_t), \\
    \mathbf{s}_{t+1} &= \mathrm{GRU}_\theta(\mathbf{g}_t).\\
    \end{aligned}
    \label{eq:decode}
\end{equation}

The perception module, $f_\theta$, the inference module, $f_\phi$, and the dynamics modules, $f_\psi$, are shared among all episodes in the dataset consisting of various causal graphs with different discrete and continuous hidden confounders, which enables one-shot adaptation to an unseen graph at test time and allows counterfactual predictions by intervening on the identified graph and rolling into the future using the dynamics module.
The inference module and the dynamics module are trained together by minimizing the following objective:
\begin{equation}
    \min_{\phi, \psi} \sum_{m} \sum_{t}\mathcal{L}(\tilde{\mathcal{V}}^{t+1}, f_\psi(\tilde{\mathcal{V}}^{1:t}, \tilde{\mathcal{E}})).
    \label{eqn:objective}
\end{equation}


\begin{algorithm}[H]
    \DontPrintSemicolon
    \KwInput{Training set with samples from the observation $S = [s_1, ..., s_T]$, initial parameters $\phi,\theta,\beta,\psi, \mathbb{G}=( \mathcal{V}^{1:T}, \mathcal{E}^{1:T})$, batch-size $b$, total iteration number $N$}
    \For{$i=1,\dots,N$}{
    \For{$t=1,\dots,T$}{
    Calculate upward embeddings $z_u$ with bottom-up modules\\
    Learn the current and prior distributional parameters from $\mathcal{N}(0,I)$, and calculate the KL divergence\\
    Calculate downward embeddings $z_t$ with top-down modules\\
    Learn the dynamic graph at level $m$ and infer the causal effects $z_c$\\
    Use the RNN decoder to calculate the current prediction with equations in~\ref{eq:decode} and measure MSE loss
    }
    Update $\phi, \theta, \beta, \psi$ using the gradients
    }
    \KwReturn{$\phi,\theta,\beta, \psi$}
    \caption{Training procedure of \textsc{REcurrent partItioned Network} (REIN)}
    \label{algo}
\end{algorithm}

\subsection{Distributional Estimation with Temporal Variational Auto-Encoder}

With VAE handling the prediction at each time step, RNN could better model the temporal dependence over time. Formally, given a sampled sequence $S = [s_1, ..., s_T]$, VAE aims to maximize the probability sampled from the learned model distribution. At time t, RNN module $p_\phi(s_t |s_{1:t-1}, \vz_{1:t})$ predicts the current pose pt having latent variables z1:t , and conditioned on previous states $s_{1:t-1}$. We rely on a variational neural network $q_\phi(\vz_t |s_{1:t})$ to approximate the true unknown posterior distribution $p_\theta(\vz_t |s_{1:t})$. This way, the objective of maximizing the data likelihood over the real sequence could be achieved as the following variational lower bound:

\begin{equation}
\begin{aligned}
    \log p_{\theta}(s)=& \log \int_{\vz} p_{\theta}(s\mid\vz) p(\vz) \geq \mathbb{E}_{q_{\phi}(\vz \mid s)} \log p_{\theta}(s \mid \vz) -D_{KL}\left(q_{\phi}(\vz \mid s) \| p(\vz)\right) \\
    =& \sum_{t}\left[\mathbb{E}_{q_{\phi}\left(\vz_{t} \mid s_{1: t}\right)} \log p_{\theta}\left(s_{t} \mid s_{1: t-1}, \vz_{1: t}\right) - D_{KL}\left(q_{\phi}\left(\vz_{t} \mid s_{1: t}\right) \| p\left(\vz_{t}\right)\right)\right]. 
\end{aligned}
\end{equation}


Here, the first term of the lower bounding function encourages the generated sample to be sufficiently close to the real sample; the second term penalizes the KL-divergence between prior and posterior distributions. A simple form of the prior $p(\vz_t)$ in VAE is a Gaussian with unitary variance, $N(0, I)$. In practice, however, the prior distribution often vary with time. Take motions from the walk category for example, sometimes the pose variance could be small (\eg, roaming); sometimes could be large (\eg, directional change of directions or velocity). The prior thus needs to be flexible enough to accommodate these variations. Intuitively, the prior of present time could be guessed given the context of previous time steps. We parameterize the prior with a variational neural network $p_\phi(\vz_t |s_{1:t-1})$ conditioned on previous steps $s_{1:t-1}$. Therefore, the variational lower bound of the sequence could be re-written as
\begin{equation}
\begin{aligned}
    \log p_{\theta}(s) \geq & \sum_{t}\left[\mathbb{E}_{q_{\phi}\left(\vz_{t} \mid s_{1: t}\right)} \log p_{\theta}\left(s_{t} \mid s_{1: t-1}, \vz_{1: t}\right)\right.\\
    &\left.-D_{KL}\left(q_{\phi}\left(\vz_{t} \mid s_{1: t}\right) \| p_{\psi}\left(\vz_t \mid s_{1: t-1}\right)\right)\right].
\end{aligned}
\end{equation}
The constraint between prior and posterior distribution further encourages temporal consistency.



\section{Experiments}\label{sec:experiment}

\textbf{Implementation} Our encoder implementation uses MLPs or GNNs with attentive pooling as our message passing function. For the decoder, fully-connected networks or a recurrent decoder are applied. All modules are built in Pytorch\footnote{\url{https://pytorch.org/}} and trained in an end-to-end style. Optimization was performed using the Adam~\cite{kingma2014adam}. We perform experiments in simple physical systems and human motion synthesis. The level number of division is 2 in simulating physical systems with limited constituents and 3 levels are used in modeling the interaction of joints in the human motion sequences. We provide more details about the data sources in the appendix.

\subsection{Physical Simulations}
\begin{figure}[t]
    \centering
    \includegraphics[width=\hsize]{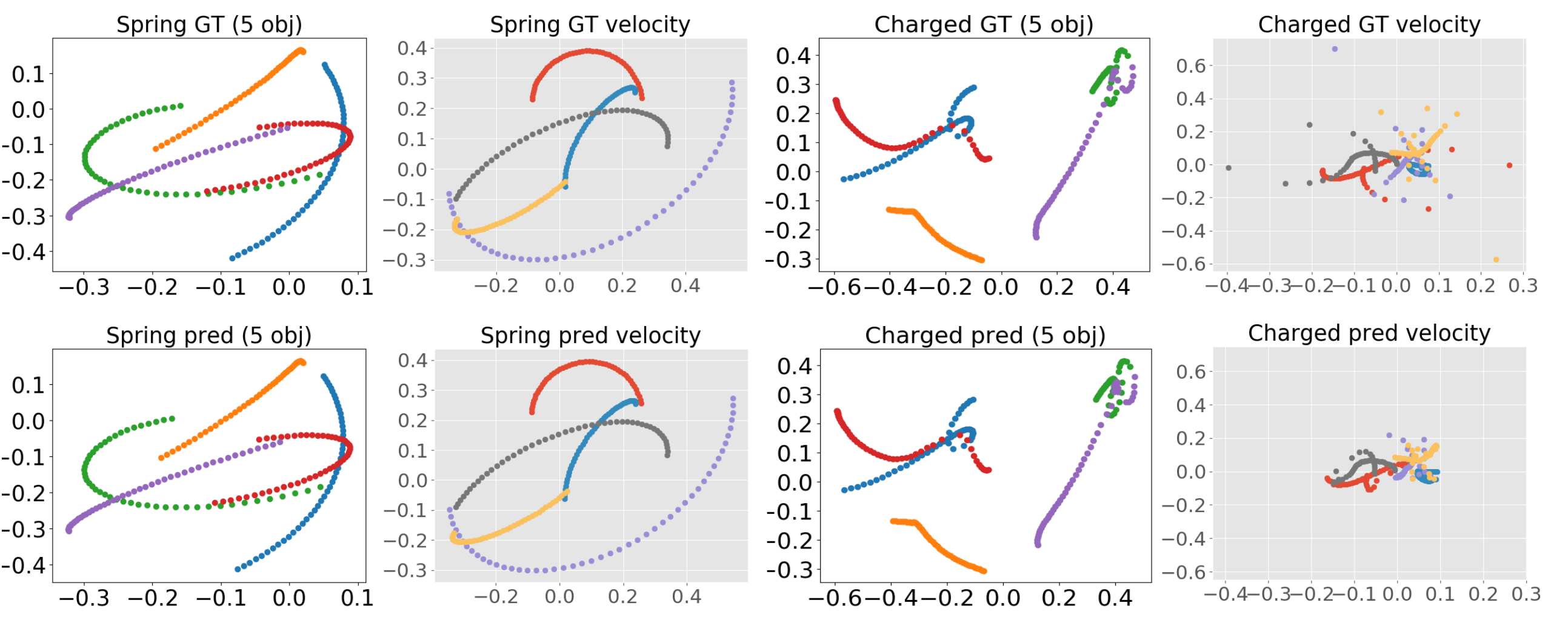}
    \vspace{3mm}
    \caption{Comparison between the GT and predicted trajectories. The encoder are trained on the first time steps, then predict the following unseen time steps. The estimated trajectories and velocities are well matched with the ground truth.}
    \label{fig:springs}
\end{figure}

We first carry out experiments simulating three different physical systems: particles connected by springs, charged particles and multi-ball interaction. These settings allow us to attempt to learn the dynamics and interactions when the interactions are known and elements can be easily measured. Such systems, manipulated by simple rules, can perform complex dynamics. Example trajectories can be seen in Fig.~\ref{fig:springs}.

\textbf{Springs \& particles} 50k training examples, and 10k validation and test examples are generated and sampled for all tasks. 
We note that the simulations are differentiable and so we can use it as a ground-truth decoder to train the encoder. The charged particles simulation, however, suffers from instability which led to some performance issues when calculating gradients.

\begin{table*}[t!]
\setlength{\tabcolsep}{3.5pt}
\centering
\caption{Mean squared error (MSE) in predicting future states for simulations with 5 interacting objects. $^{\star}$ indicates model (MLP) learned with ground truth graph.}
\vspace{2mm}
\begin{tabular}{l cccc c cccc}
    \toprule
    & \multicolumn{4}{c}{\textbf{Springs}} &  & \multicolumn{4}{c}{\textbf{Charged}} \\ 
    
    \midrule
    Steps    & 1 & 10 & 20 & 50 &   & 1 & 10 & 20 & 50 \\ 
    \midrule
    GT$^{\star}$  & \ee{1.38}{e-11} & \ee{1.14}{e-9} & \ee{6.63}{e-6} & \ee{4.25}{e-5} &
                             & \ee{1.06}{e-3} & \ee{3.20}{e-3} & \ee{5.28}{e-3}  & \ee{9.47}{e-3} \\
    \hdashline
    Static   & \ee{7.93}{e-5} & \ee{7.59}{e-3} & \ee{2.82}{e-2} & \ee{4.63}{e-2} &
    & \ee{5.09}{e-3} & \ee{2.26}{e-2} & \ee{5.42}{e-2}  & \ee{8.85}{e-2}  \\
    LSTM     & \ee{4.13}{e-8} & \ee{2.19}{e-5} & \ee{7.02}{e-4} & \ee{6.73}{e-3} &
    & \ee{1.68}{e-3} & \ee{6.45}{e-3} & \ee{1.49}{e-2}  & \ee{4.61}{e-2}  \\
    NRI~\cite{kipf2018neural}    & \ee{3.12}{e-8} & \ee{3.29}{e-6} & \ee{2.13}{e-5} &\ee{1.30}{e-4} &
    & \eeb{1.05}{e-3} & \ee{3.21}{e-3} & \ee{7.06}{e-3}  & \ee{2.18}{e-2}  \\
    \midrule
    \textbf{REIN}  & \eeb{2.69}{e-9} & \eeb{5.32}{e-7} & \eeb{7.06}{e-6}  & \eeb{8.34}{e-5} &
          & \ee{1.24}{e-3} & \eeb{1.63}{e-3} & \eeb{6.71}{e-3}  & \eeb{1.71}{e-2} \\
    
    \bottomrule
\end{tabular}
\vspace{3mm}
\label{tab:physics}
\end{table*}

\begin{wraptable}{r}{7cm}
\setlength{\tabcolsep}{9.9pt}
  \caption{Accuracy measured for multi-ball system.}
  \begin{tabular}{l cc }
    \toprule
                    & Accuracy-edge & MSE-location  \\   
    \midrule
            Static    &  \et{0.736}{.021}& \et{8.437}{e-2}\\
            LSTM    &  \et{0.873}{.014}& \et{3.264}{e-3}\\
            NRI    &  \et{0.925}{.019}& \et{6.853}{e-5}\\
    \midrule
           
            \textbf{REIN}    &  \etb{0.947}{.012}  & \etb{2.309}{e-5} \\
    \bottomrule
  \end{tabular}
  \label{tab:physic-scm}
\end{wraptable}


\textbf{Multi-ball system} In the multi-ball interaction, there are 5 balls of different colors moving around. At the beginning of each episode, we sample the invisible physical relations between each pair of balls independently, giving us the ground truth that is fixed throughout the episode. For each pair of balls, there is a one-third probability that they are not connected or linked by a rigid rod or a spring. We also sample the continuous parameters for each existing edge and fix them within the episode, e.g., the length of the rigid relation or the rest length of the spring.

We evaluate the proposed REIN on all three simulated physical systems and compared our performance, both in future state prediction and in the accuracy of estimating the causal connection between nodes in the graph with no supervision. Additionally, we also compare to other baseline methods: MLPs model with the ground-truth simulation decoder and two correlation-based baselines, Static and LSTM.
In order to have a fair comparison, we generate longer test trajectories in steps (1, 10, 20, 50) and only evaluate on the last part unseen by the encoder. We first input in the LSTM a number of samples, and then predict the location in next time steps. We show mean squared error (MSE) results in Table~\ref{tab:physics}. It is observed that our results are better than others especially for long term prediction and are also comparable to the model trained with ground truth graph connections. We also show some generated trajectories in Fig.~\ref{fig:springs} compared to the ground-truth trajectories.

For the multi-ball system, we report the accuracy of predicted edges and also the MSE loss of final predictions in Table~\ref{tab:physic-scm}. After the mapping, we evaluate the model in predicting the continuous confounder, by computing its correlation with the ground truth physical parameters. With more samples observed, the classification accuracy increases, and the uncertainty decreases, which 
verifies that with more observations from the environment, the model can better estimate the dynamic formulation that governs the behavior of the system. It is proved that our model outperforms the baselines, which also indicates the importance of causal relationship inference.

\subsection{Human 3D Motion Synthesis}

Motion sequences can be regarded as the coordination of all the joints in the skeleton which provides a suitable bedrock for the evaluation of the motivation of REIN.
The datasets adopted in our experiments possess various action categories, where each type contains a considerable amount of motions in diverse styles, recorded with delicate annotations. 
In the datasets, skeletons are all composed of 5 kinematic chains: \{spine, left arm, right arm, left leg, right leg\}, and the root joint is located at the end of the hipbone. Our decoder first produces the Lie algebraic parameters and root translation via linear layers and graph networks, then obtains the 3D positions of the skeleton via the forward kinematics. The motion states at each time step include the 3D location of each joint as well as the velocity by calculating the difference value with the previous step.

\textbf{Datasets} \textit{NTU-RGB-D}~\cite{liu2019ntu} originally contains 120 action types of 106 subjects. Its pose representation (3D joint positions) is from MS Kinect readout, which is known to unreliable and temporally inconsistent. It is sufficient here to be perceptually natural and realistic. We adopted a subset of 13 distinct actions including e.g. \textit{cheer up, pick up, salute}, constituting 3,900 motion clips.
\textit{HumanAct12}~\cite{chuan2020a2m} is the in-house dataset, consisting of 1,191 motion clips and 90,099 frames in total, with hierarchical action type annotations. To be specific, all motions are organized into 12 action categories, including \textit{warm up}, \textit{lift dumbbell}, and 34 subcategories including \textit{jump handsup}, \textit{eat with left hand}) which offer more detailed commands of motion. Compared to NTU-RGB-D, the 3D position annotations are more accurate, and the pose sequences are more stable. 

\textbf{Evaluation metrics} \textit{Frechet Inception Distance} (FID) is an important metric widely used to evaluate the overall quality of generated motions. Features are extracted from generated motions and real motions by sampling with replacement from the test set. Then FID is calculated between the feature distribution of generated motions vs. that of the real motions.
\textit{Recognition Accuracy} is calculated with a pre-trained RNN action recognition classifier to classify the motions, and calculate the overall recognition accuracy. The recognition accuracy indicates the correlation of the motion and its action type.
\textit{Diversity} measures the variance of the generated motions across all action categories. From a set of all generated motions from various action types, two subsets of the same size $d$ are randomly sampled. Their respective sets of motion feature vectors $\{\mathbf{v}_1,...,\mathbf{v}_{d}\}$ and $\{\mathbf{v}_1',...,\mathbf{v}_{d}'\}$ are extracted. The diversity of this set of motions is defined as follows where $d$ is set to 200 as in~\cite{chuan2020a2m}:
\begin{equation}
    \mathrm{Diversity} = \frac{1}{d}\sum_{i=1}^{d}\parallel \mathbf{v}_i-\mathbf{v}_i' \parallel_2.
\end{equation}

\begin{table*}[t]
\setlength{\tabcolsep}{3.8pt}
  \caption{Performance evaluation on HumanAct12 and NTU-RGB-D. $^{\star}$ indicates model learned with ground truth human kinetic chain.}
  \begin{tabular}{l ccc c ccc }
    \toprule
    \multirow{2}{*}{} & \multicolumn{3}{c}{\textbf{HumanAct12}} & & \multicolumn{3}{c}{\textbf{NTU-RGB-D}} \\
    \cline{2-8}
                    & FID$\downarrow$ & Accuracy & Diversity  & 
                    & FID$\downarrow$ & Accuracy & Diversity \\   
    \midrule
            GT motions    & \et{0.092}{.007}  &  \et{0.997}{.001}& \et{6.853}{.053}& & \et{0.031}{.004}  &  \et{0.999}{.001}& \et{7.108}{.048} \\
    \midrule
            CondGRU   & \et{40.61}{.144}  &  \et{0.080}{.002} & \et{2.381}{.020}& &  \et{28.31}{.138}  &  \et{0.078}{.001}& \et{3.663}{.024} \\
            Two-stage    & \et{10.48}{.089} &  \et{0.421}{.006} & \et{5.960}{.049}& & \et{13.86}{.091}  &  \et{0.202}{.003}& \et{5.328}{.039} \\
            Act-MoCo   & \et{5.610}{.113}  &  \et{0.793}{.004} & \et{6.752}{.071}& & \et{2.723}{.019}  &  \etb{0.997}{.001}& \et{6.920}{.061} \\
            t-VAE~\cite{chuan2020a2m}    & \et{2.458}{.079}  &  \et{0.923}{.002} & \et{7.032}{.038}& & \et{0.330}{.008} &  \et{0.949}{.001}& \et{7.065}{.043} \\
    \midrule
            REIN$^{\star}$    & \et{0.392}{.034}  &  \et{0.933}{.005}   & \et{6.417}{.046} & 
                              & \et{0.340}{.027}  &  \et{0.832}{.004}   & \et{6.514}{.049} \\
            \textbf{REIN}    & \etb{0.270}{.013}  &  \etb{0.947}{.002}  & \etb{7.309}{.038} & 
                             & \etb{0.130}{.011}  &  \et{0.949}{.001}   & \etb{7.723}{.043} \\
    \bottomrule
  \end{tabular}
  \label{tab:motion}
\end{table*}

\begin{figure}[t]
    \centering
    \vspace{3mm}
    \includegraphics[width=\hsize]{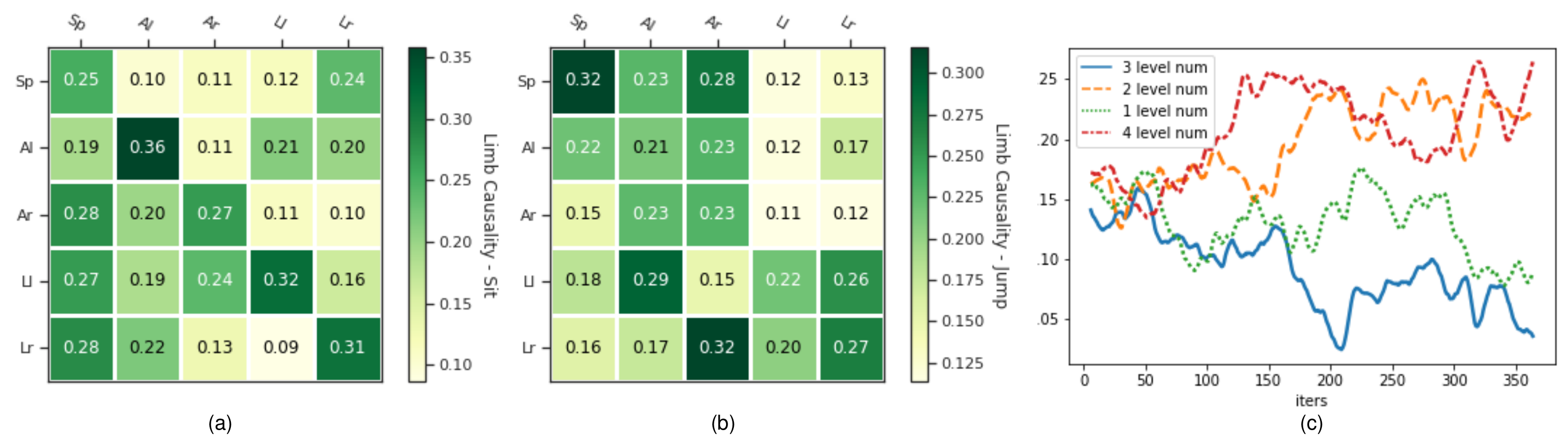}
    \caption{Left: Examples of predicted scores for edges in the causal graph for peer propagation at the limb level (spine, left arm, right arm, left leg, right leg) averaged over time (a) for action ``Sit'' and (b) for ``Jump''. (c) shows the loss during training with hierarchical level number from 1 to 4.}
    \label{fig:motion_stats}
\end{figure}


With results reported in Table~\ref{tab:motion}, it proves that our method qualitatively overwhelms other methods not only in the FID metric and categorical accuracy but also in the diversity of the generated poses. 
To evaluate the effectiveness of our structure-aware modules, we replaced the graph convolution with a standard 1D convolution with full support over the channels axis, and the skeletal pooling and unpooling are discarded and upsampling and downsampling is performed only on the temporal axis. 
Hierarchical partitioning the overall structure into functional units as shown in Fig.~\ref{fig:motion_stats}. (c) shows the total loss during training where we can find that with 3 levels for the partitioning, REIN can be better optimized (see Appendix for more details). 
In Fig.~\ref{fig:motion_causality}, we show the learn causal connection of sub-parts in human skeleton. In order to support the structure of a shared latent space, we simply modify the number of output and input channels in each encoder and decoder, respectively, such that both auto-encoders share the same latent space size, which equals to the one in the original distribution. 

\begin{figure}[t]
    \centering
    \includegraphics[width=\hsize]{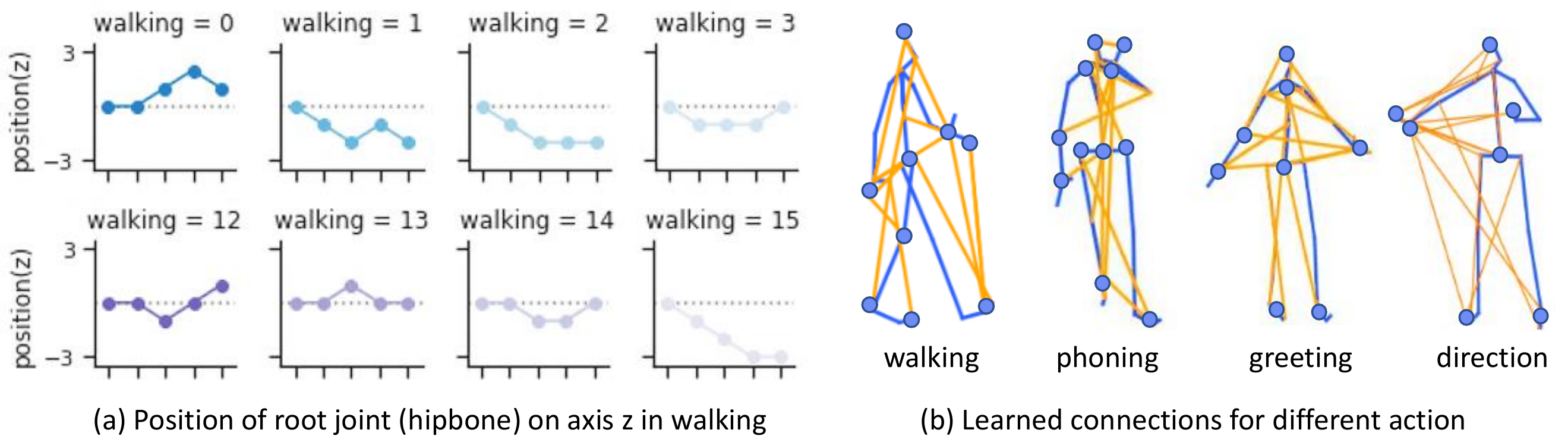}
    \caption{The z-axis position of the root joint at the beginning steps and middle period which has high correlation with the movement of two feet. On the right, we show the causality learned among the joints with several examples of different action types.}
    \label{fig:motion_causality}
\end{figure}




\subsection{Ablation and Further Discussion}~\label{sec:discussion}

Here we would like to share some reflection on the design details of the proposed architecture. With recurrent partitioned neurons, the system can be modeled with a part-whole hierarchy with predefined levels. We conduct ablation study on HumanAct12 to evaluate the effectiveness of each module in REIN with results reported in Table~\ref{tab:motion-abl}. With a single branch in upward or downward, REIN is not able to take full control of the dynamics generated and performs barely satisfactory. We also compared the peer communication layer to verify the learned connections compared to randomly connected ones. It profits from learning representations from the high global level into the structural sub-part levels that possess like-minded embeddings.

\begin{wraptable}{r}{8cm}
\setlength{\tabcolsep}{8.5pt}
  \caption{Ablation study of modules in REIN for motion synthesis on HumanAct12.}
  \begin{tabular}{l cc }
    \toprule
                    & Accuracy & Diversity  \\   
    \midrule
            GT motions     &  \et{0.997}{.001}& \et{6.853}{.053}\\
    \midrule
            REIN (upward)     &  \et{0.762}{.035}  & \et{3.697}{.085} \\
            REIN (downward)     &  \et{0.850}{.028}  & \et{4.386}{.062} \\
    \hdashline
            REIN (p-random)   &  \et{0.904}{.007}  & \et{6.263}{.049} \\
            REIN (p-learned)   &  \et{0.904}{.007}  & \et{6.263}{.049} \\
    \midrule
            \textbf{REIN}     &  \etb{0.947}{.002}  & \etb{7.309}{.038} \\
    \bottomrule
  \end{tabular}
  \label{tab:motion-abl}
\end{wraptable}

\textit{First, how fine-grained should the inner components be divided?} 
In our experiment, we leverage the common-sense understanding of the physical structure and combination. For example, joints in the spine are grouped together and so do joints in the arms and legs. More detailly, each pair of two adjacent joints in a bone can be taken as a group and perform rigidly together. However, limitations could be foreseen when the physical structure is hard to observe or incompatible with the underlying dynamic hierarchy.

\textit{Secondly, how to efficiently maintain the most valuable information during the propagation?} One valuable question is whether the bottom-up and top-down channels can be shared across levels as well as across locations. This would not work for lower levels if the extraction of features are not compact or lost during the transmit. An advantage in message propagating across levels exists in that the representation at different level and steps would be aligned to keep the consistency and distributional attributes. This would make it easier to seek for the gold in the fine details as well as having a sense of what is happening at the high levels.

\section{Conclusions}\label{sec:conclusion}

This paper carries out a study on how to simulate the dynamic process by representing the system in a hierarchical structure and inferring the componential embeddings with structural causality. In the experiments, our method first establishes a structured graph-based representation from observation, identifies the causal relationships between different variables, and makes predictions based on the bottom-up, top-down and peer communication messages. Instead of relying directly on the full supervision of the inner relationship, REIN learns to discover the dependency structures and model the causal mechanisms dynamically, which can better cope with the simulation of more complex and versatile dynamic processes.

{\small
\bibliographystyle{ieee_fullname}
\bibliography{reference}

\begin{thebibliography}{10}\itemsep=-1pt

\bibitem{bateman1997does}
Ian Bateman, Alistair Munro, Bruce Rhodes, Chris Starmer, and Robert Sugden.
\newblock Does part--whole bias exist? an experimental investigation.
\newblock {\em The Economic Journal}, 107(441):322--332, 1997.

\bibitem{ellner2011dynamic}
Stephen~P Ellner and John Guckenheimer.
\newblock {\em Dynamic models in biology}.
\newblock Princeton University Press, 2011.

\bibitem{girju2003learning}
Roxana Girju, Adriana Badulescu, and Dan Moldovan.
\newblock Learning semantic constraints for the automatic discovery of
  part-whole relations.
\newblock In {\em Proceedings of the 2003 Human Language Technology Conference
  of the North American Chapter of the Association for Computational
  Linguistics}, pages 80--87, 2003.

\bibitem{goyal2020dyngraph2vec}
Palash Goyal, Sujit~Rokka Chhetri, and Arquimedes Canedo.
\newblock dyngraph2vec: Capturing network dynamics using dynamic graph
  representation learning.
\newblock {\em Knowledge-Based Systems}, 187:104816, 2020.

\bibitem{granger1980testing}
Clive~WJ Granger.
\newblock Testing for causality: a personal viewpoint.
\newblock {\em Journal of Economic Dynamics and control}, 2:329--352, 1980.

\bibitem{chuan2020a2m}
Chuan Guo, Xinxin Zuo, Sen Wang, Shihao Zou, Qingyao Sun, Annan Deng, Minglun
  Gong, and Li Cheng.
\newblock Action2motion: Conditioned generation of 3d human motions.
\newblock In {\em Proceedings of the 28th ACM International Conference on
  Multimedia (MM '20)}, 2020.

\bibitem{hajiramezanali2019variational}
Ehsan Hajiramezanali, Arman Hasanzadeh, Nick Duffield, Krishna~R Narayanan,
  Mingyuan Zhou, and Xiaoning Qian.
\newblock Variational graph recurrent neural networks.
\newblock {\em arXiv preprint arXiv:1908.09710}, 2019.

\bibitem{higgins2016beta}
Irina Higgins, Loic Matthey, Arka Pal, Christopher Burgess, Xavier Glorot,
  Matthew Botvinick, Shakir Mohamed, and Alexander Lerchner.
\newblock beta-vae: Learning basic visual concepts with a constrained
  variational framework.
\newblock 2016.

\bibitem{khatib2009robotics}
Oussama Khatib, Emel Demircan, Vincent De~Sapio, Luis Sentis, Thor Besier, and
  Scott Delp.
\newblock Robotics-based synthesis of human motion.
\newblock {\em Journal of Physiology-Paris}, 103(3-5):211--219, 2009.

\bibitem{kingma2014adam}
Diederik~P Kingma and Jimmy Ba.
\newblock Adam: A method for stochastic optimization.
\newblock In {\em International Conference on Learning Representations (ICLR)},
  2015.

\bibitem{kipf2018neural}
Thomas Kipf, Ethan Fetaya, Kuan-Chieh Wang, Max Welling, and Richard Zemel.
\newblock Neural relational inference for interacting systems.
\newblock In {\em International Conference on Machine Learning}, pages
  2688--2697. PMLR, 2018.

\bibitem{kumar2017variational}
Abhishek Kumar, Prasanna Sattigeri, and Avinash Balakrishnan.
\newblock Variational inference of disentangled latent concepts from unlabeled
  observations.
\newblock {\em arXiv preprint arXiv:1711.00848}, 2017.

\bibitem{kyono2020castle}
Trent Kyono, Yao Zhang, and Mihaela van~der Schaar.
\newblock Castle: Regularization via auxiliary causal graph discovery.
\newblock {\em arXiv preprint arXiv:2009.13180}, 2020.

\bibitem{liu2019ntu}
Jun Liu, Amir Shahroudy, Mauricio~Lisboa Perez, Gang Wang, Ling-Yu Duan, and
  Alex~Kot Chichung.
\newblock Ntu rgb+d 120: A large-scale benchmark for 3d human activity
  understanding.
\newblock {\em IEEE transactions on pattern analysis and machine intelligence},
  2019.

\bibitem{locatello2019challenging}
Francesco Locatello, Stefan Bauer, Mario Lucic, Gunnar Raetsch, Sylvain Gelly,
  Bernhard Sch{\"o}lkopf, and Olivier Bachem.
\newblock Challenging common assumptions in the unsupervised learning of
  disentangled representations.
\newblock In {\em international conference on machine learning}, pages
  4114--4124. PMLR, 2019.

\bibitem{luo2020dynamic}
Wenjuan Luo, Han Zhang, Xiaodi Yang, Lin Bo, Xiaoqing Yang, Zang Li, Xiaohu
  Qie, and Jieping Ye.
\newblock Dynamic heterogeneous graph neural network for real-time event
  prediction.
\newblock In {\em Proceedings of the 26th ACM SIGKDD International Conference
  on Knowledge Discovery \& Data Mining}, pages 3213--3223, 2020.

\bibitem{rezende2014stochastic}
Danilo~Jimenez Rezende, Shakir Mohamed, and Daan Wierstra.
\newblock Stochastic backpropagation and approximate inference in deep
  generative models.
\newblock In {\em International conference on machine learning}, pages
  1278--1286. PMLR, 2014.

\bibitem{seo2018structured}
Youngjoo Seo, Micha{\"e}l Defferrard, Pierre Vandergheynst, and Xavier Bresson.
\newblock Structured sequence modeling with graph convolutional recurrent
  networks.
\newblock In {\em International Conference on Neural Information Processing},
  pages 362--373. Springer, 2018.

\bibitem{song2019session}
Weiping Song, Zhiping Xiao, Yifan Wang, Laurent Charlin, Ming Zhang, and Jian
  Tang.
\newblock Session-based social recommendation via dynamic graph attention
  networks.
\newblock In {\em Proceedings of the Twelfth ACM International Conference on
  Web Search and Data Mining}, pages 555--563, 2019.

\bibitem{Sugihara496}
George Sugihara, Robert May, Hao Ye, Chih-hao Hsieh, Ethan Deyle, Michael
  Fogarty, and Stephan Munch.
\newblock Detecting causality in complex ecosystems.
\newblock {\em Science}, 338(6106):496--500, 2012.

\bibitem{villaverde2016structural}
Alejandro~F Villaverde, Antonio Barreiro, and Antonis Papachristodoulou.
\newblock Structural identifiability of dynamic systems biology models.
\newblock {\em PLoS computational biology}, 12(10):e1005153, 2016.

\bibitem{winston1987taxonomy}
Morton~E Winston, Roger Chaffin, and Douglas Herrmann.
\newblock A taxonomy of part-whole relations.
\newblock {\em Cognitive science}, 11(4):417--444, 1987.

\bibitem{wu2019graph}
Zonghan Wu, Shirui Pan, Guodong Long, Jing Jiang, and Chengqi Zhang.
\newblock Graph wavenet for deep spatial-temporal graph modeling.
\newblock {\em arXiv preprint arXiv:1906.00121}, 2019.

\bibitem{yang2020causalvae}
Mengyue Yang, Furui Liu, Zhitang Chen, Xinwei Shen, Jianye Hao, and Jun Wang.
\newblock Causalvae: Structured causal disentanglement in variational
  autoencoder.
\newblock {\em arXiv preprint arXiv:2004.08697}, 2020.

\bibitem{ye2015distinguishing}
Hao Ye, Ethan~R Deyle, Luis~J Gilarranz, and George Sugihara.
\newblock Distinguishing time-delayed causal interactions using convergent
  cross mapping.
\newblock {\em Scientific reports}, 5(1):1--9, 2015.

\bibitem{young1988variable}
K-KD Young.
\newblock A variable structure model following control design for robotics
  applications.
\newblock {\em IEEE Journal on Robotics and Automation}, 4(5):556--561, 1988.

\bibitem{zhang2017random}
Xiao Zhang, Cristopher Moore, and Mark~EJ Newman.
\newblock Random graph models for dynamic networks.
\newblock {\em The European Physical Journal B}, 90(10):1--14, 2017.

\bibitem{vzlajpah2008simulation}
Leon {\v{Z}}lajpah.
\newblock Simulation in robotics.
\newblock {\em Mathematics and Computers in Simulation}, 79(4):879--897, 2008.

\end{thebibliography}
}

\end{document}